\documentclass[sigconf]{acmart}
\AtBeginDocument{%
  }

\usepackage{cuted}
\usepackage{caption}
\usepackage{multirow}
\usepackage{makecell}
\usepackage{tabularx,booktabs,caption,ragged2e}
\usepackage{amsmath,amsfonts}
\usepackage{amsthm}
\usepackage{array}

\usepackage{graphicx,caption}
\usepackage{longtable}

\usepackage{algorithm}
\usepackage{algorithmic}

\usepackage{float} 
\newcolumntype{Y}{>{\RaggedRight\arraybackslash}X}

\newcommand{\numstd}[2]{
    #1{\scriptsize$\pm$#2}
}

\setcopyright{acmlicensed}
\copyrightyear{2026}
\acmYear{2026}
\setcopyright{cc}
\setcctype{by}
\acmConference[KDD '26]{Proceedings of the 32nd ACM SIGKDD Conference on Knowledge Discovery and Data Mining V.2}{August 09--13, 2026}{Jeju Island, Republic of Korea}
\acmBooktitle{Proceedings of the 32nd ACM SIGKDD Conference on Knowledge Discovery and Data Mining V.2 (KDD '26), August 09--13, 2026, Jeju Island, Republic of Korea}
\acmDOI{10.1145/3770855.3818939}
\acmISBN{979-8-4007-2259-2/2026/08}

\begin{document}

\title[CausalMoE: A Billion-Scale Multimodal Foundation Model for Granger Causal Discovery]{CausalMoE: A Billion-Scale Multimodal Foundation Model for Granger Causal Discovery with Pattern-Routed Heterogeneous Experts}

\author{Bo Liu}
\affiliation{%
  \institution{State Key Laboratory of General Artificial Intelligence, School of Intelligence Science and Technology, Peking University}
  \city{Beijing}
  \country{China}}
\email{liubo2022@stu.pku.edu.cn}

\author{Di Dai}
\affiliation{%
  \institution{State Key Laboratory of General Artificial Intelligence, School of Intelligence Science and Technology, Peking University}
  \city{Beijing}
  \country{China}}
\email{didai@stu.pku.edu.cn}
\authornote{Corresponding Authors.}

\author{Jingwei Liu}
\author{Jiarui Jin}
\affiliation{%
  \institution{State Key Laboratory of General Artificial Intelligence, School of Intelligence Science and Technology, Peking University}
  \city{Beijing}
  \country{China}}

\author{Xiaocheng Fang}
\author{Guangkun Nie}
\affiliation{%
  \institution{State Key Laboratory of General Artificial Intelligence, School of Intelligence Science and Technology, Peking University}
  \city{Beijing}
  \country{China}}

\author{Hongyan Li}
\affiliation{%
  \institution{State Key Laboratory of General Artificial Intelligence, School of Intelligence Science and Technology, Peking University}
  \city{Beijing}
  \country{China}}
\email{leehy@pku.edu.cn}

\author{Shenda Hong}
\affiliation{%
  \institution{National Institute of Health Data Science, and Institute for Artificial Intelligence, Peking University}
  \city{Beijing}
  \country{China}}
\email{hongshenda@pku.edu.cn}
\authornotemark[1]

\renewcommand{\shortauthors}{Bo Liu et al.}

\begin{abstract}
Granger Causal Discovery (GCD) is fundamental for analyzing temporal dependencies in complex systems. However, existing neural GCD methods predominantly rely on a "one-size-fits-all" paradigm, struggling to capture distribution shifts and dynamic regime changes inherent in real-world time series. This often leads to entangled representations and spurious causal graphs. In this paper, we propose CausalMoE, a billion-scale multimodal Granger causal foundation model that explicitly models patch-level heterogeneity. CausalMoE introduces a Pattern-Routed Mixture of Heterogeneous Experts, which dynamically identifies latent temporal patterns and routes patches to specialized domain experts, effectively decoupling regime-specific mechanisms from shared dynamics. To ensure interpretable graph recovery, we design a Causality-Aware Self-Attention mechanism operating across variables, yielding sparse Granger causal graphs via proximal optimization. Furthermore, CausalMoE is the first to integrate LLMs and VLMs to align numerical signals with textual and visual priors, regularizing causal estimation in complex scenarios. Extensive experiments demonstrate that CausalMoE establishes a new state-of-the-art on fully supervised benchmarks, while effectively generalizing to few-shot settings where traditional methods fail.
\end{abstract}



\maketitle

\section{Introduction}

Granger Causality (GC) \cite{GC} has long served as a fundamental paradigm for causal discovery in Time Series (TS), owing to its prediction-based formulation and interpretability. With the advent of deep learning, neural Granger causal discovery (GCD) \cite{NGC} has substantially extended classical GC to nonlinear and high-dimensional settings \cite{CUTS}, leveraging the expressivity of Transformers \cite{CausalKGR} or generative models \cite{CR-VAE,DiffuGC} to capture nonlinear dynamics. These advances have enabled GC in a wide range of real-world scenarios, including geography \cite{causalrivers}, genetics \cite{singh2022granger}, and biology \cite{yu2023deep}.

\begin{figure}
    \centering
    \includegraphics[width=\linewidth]{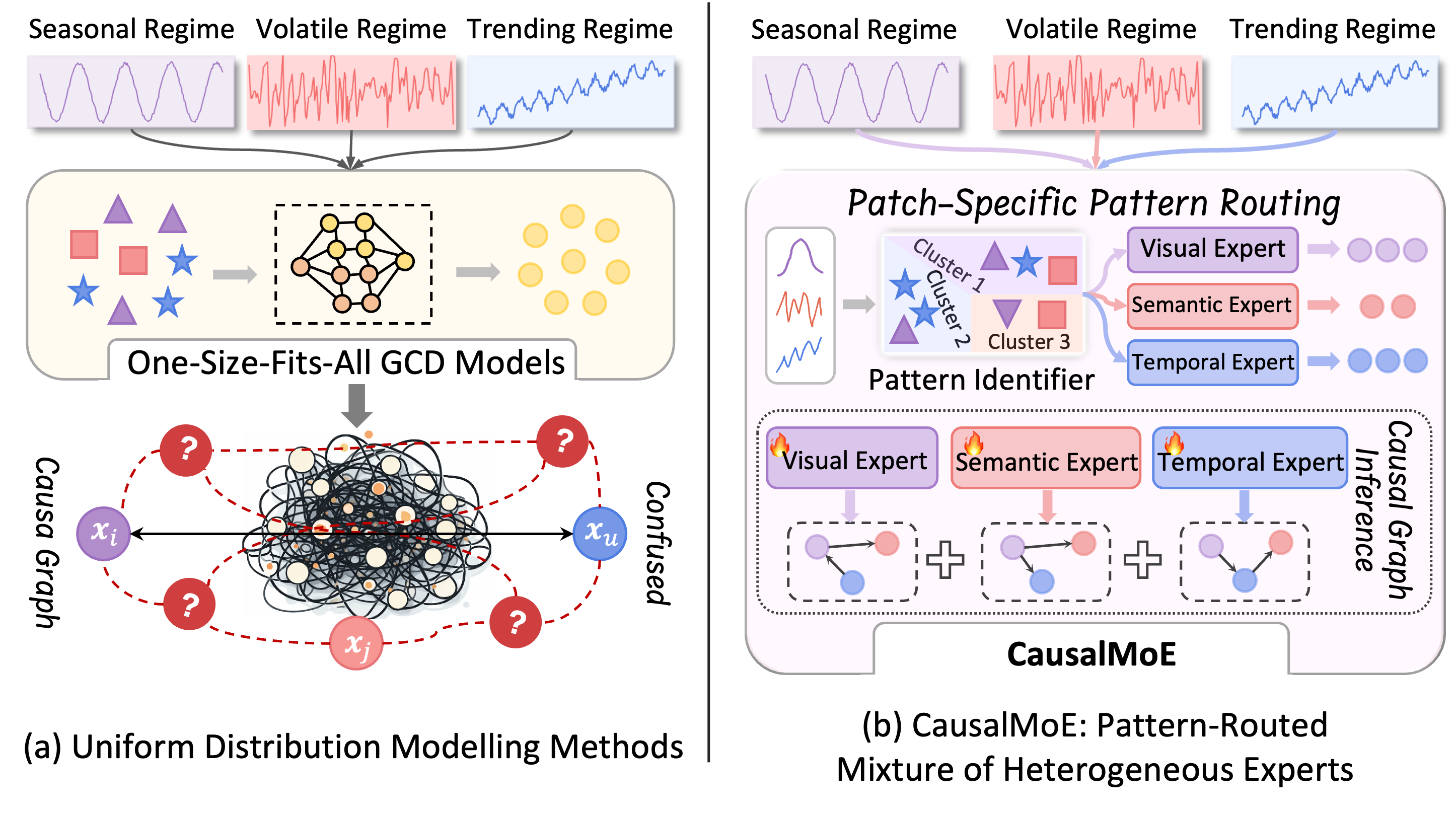}
    \caption{Addressing the one-size-fits-all limitation in GCD. 
(a) Traditional methods employ the UDM to conflate temporal regimes, leading to spurious and entangled causal graphs. (b) CausalMoE performs pattern-specific routing to heterogeneous experts, enabling regime-aware causal modeling.}

    \label{fig:motivation}
\end{figure}

Despite these advances, most existing GCD methods share a common implicit yet strong assumption: the Uniform Distribution Modeling (UDM) strategy \cite{Pattern-Specific}. The UDM posits a homogeneous data-generating process, treating all temporal segments as samples from a single underlying data-generating distribution \cite{CUTS, IGC}. However, real-world TS frequently exhibit distribution shifts and regime changes, even at the granularity of short temporal patches \cite{Pattern-Specific, DUET}. Crucially, shifts may reflect changes in underlying causality rather than noise to be normalized away. As illustrated in Fig. \ref{fig:motivation}, when heterogeneous regimes are indiscriminately aggregated by UDM-based methods, genuine causality are diluted within a "tangled" representation space, inducing spurious Granger causality driven by regime-dependent relationships. Consequently, addressing the dichotomy between static modeling assumptions and dynamic real-world heterogeneity remains a critical bottleneck for reliable GCD.

Recent Time Series Foundation Models (TSFMs) offer a promising direction by demonstrating strong few-shot forecasting performance \cite{TS-RAG,Moirai-MoE} through large-scale pretraining \cite{li2025tsfm}. However, these models address temporal heterogeneity only implicitly, optimizing for prediction rather than causal fidelity. Structural variations across regimes are often absorbed into latent representations, thereby eroding the invariances required for causal discovery. Moreover, existing GCD methods remain strictly unimodal, relying exclusively on numerical observations \cite{CUTS}. In many complex systems, numerical correlations alone are insufficient for causal identification; semantic context—such as textual descriptions, event annotations, or visual states—often acts as a critical confounder or disambiguating signal, yet remains absent from current causal discovery pipelines. As the demand for causal-discovery-as-a-service \cite{causalrivers}, a fundamental question arises: \textit{Is it possible to construct a multimodal Granger Causal Foundation Model capable of few-shot causal inference while effectively modelling heterogeneous temporal patterns?}

Answering this question drives the design of CausalMoE, a billion-scale Multimodal Granger Causal Foundation Model that departs fundamentally from one-size-fits-all causal modelling and represents a paradigm shift towards adaptive causal modelling. We propose a novel Pattern-Routed Mixture of Heterogeneous Experts (MoHE) architecture that explicitly identifies latent temporal patterns and dynamically routes each time-series patch to the most suitable domain expert. This design enables decoupling regime-specific causality from shared dynamics across experts, which are jointly mined by a causal augmenter using multiple interaction matrices. Furthermore, CausalMoE is the first to integrate Large Language Models (LLMs) and Vision-Language Models (VLMs) into the GCD loop, leveraging multimodal semantic priors to disambiguate causal links that are unidentifiable from numerical data alone. The code of CausalMoE is available at https://github.com/liubolab/CausalMoE. Our main contributions are summarized as follows:

\begin{itemize}
    \item We propose CausalMoE, a billion-scale multimodal Granger causal discovery foundation model that explicitly models patch-level temporal heterogeneity, routing patches to Pattern-Routed Mixture of Heterogeneous Experts that decouples regime-specific mechanisms from shared dynamics.
    \item We incorporate LLMs and VLMs into the causal discovery loop to align numerical signals with semantic and visual priors, recovering sparse, interpretable graphs through Causality-Aware Self-Attention with proximal optimization.
    \item Extensive experiments on synthetic and real-world benchmarks show that CausalMoE achieves state-of-the-art performance, with strong generalization under few-shot and complex temporal settings.

\end{itemize}

\section{Related Work}
\subsection{Granger Causal Discovery in Time Series}
Granger causality (GC) \cite{GC} is a standard framework for identifying temporal causal relationships by testing whether the history of one time series helps predict another. Classical VAR-based GC is limited in modeling nonlinear dependencies, motivating neural GC methods that use deep networks for more flexible causal structure learning. Existing approaches include sparse component-wise networks \cite{NGC}, forecasting-based graph learning methods \cite{JRNGC, eSRU}, generative models such as dynamic variational autoencoders \cite{CR-VAE}, and methods for irregular or incomplete time series \cite{CUTS, CUTS+}. Recent studies also explore root cause detection with abnormal exogenous signals \cite{han2025root}. Despite these advances, most neural GCD methods still rely on numerical-only inputs and assume a homogeneous data-generating process, limiting their ability to use semantic priors and adapt to regime shifts or data scarcity. CausalMoE addresses these limitations by incorporating multimodal information and modeling heterogeneous temporal patterns.

\subsection{Time Series Foundation Models}
A parallel line of work explores large pretrained models for time-series learning to mitigate data scarcity. 
LLMs have shown strong cross-domain generalization \cite{zhang2024large}, and GPT4TS \cite{GPT4TS} pioneered their use in time-series forecasting. 
Subsequent studies improve adaptation through component decomposition \cite{cao2024tempo}, statistical prompting \cite{chuang2024understanding}, textual reprogramming \cite{TimeLLM, wang2025llm}, and embedding-space alignment \cite{TEST, zhao2025stem}. 
Recent Time-Series Foundation Models further improve few-shot forecasting with MoE architectures \cite{Time-MoE,Moirai-MoE} and retrieval augmentation \cite{TS-RAG}. 
However, they mainly optimize forecasting objectives and often use homogeneous expert pools, limiting their suitability for causal discovery under heterogeneous temporal regimes. 
CausalMoE addresses these gaps by directly targeting Granger causality with heterogeneous, patch-routed experts.

\begin{figure*}
    \centering
    \includegraphics[width=\linewidth]{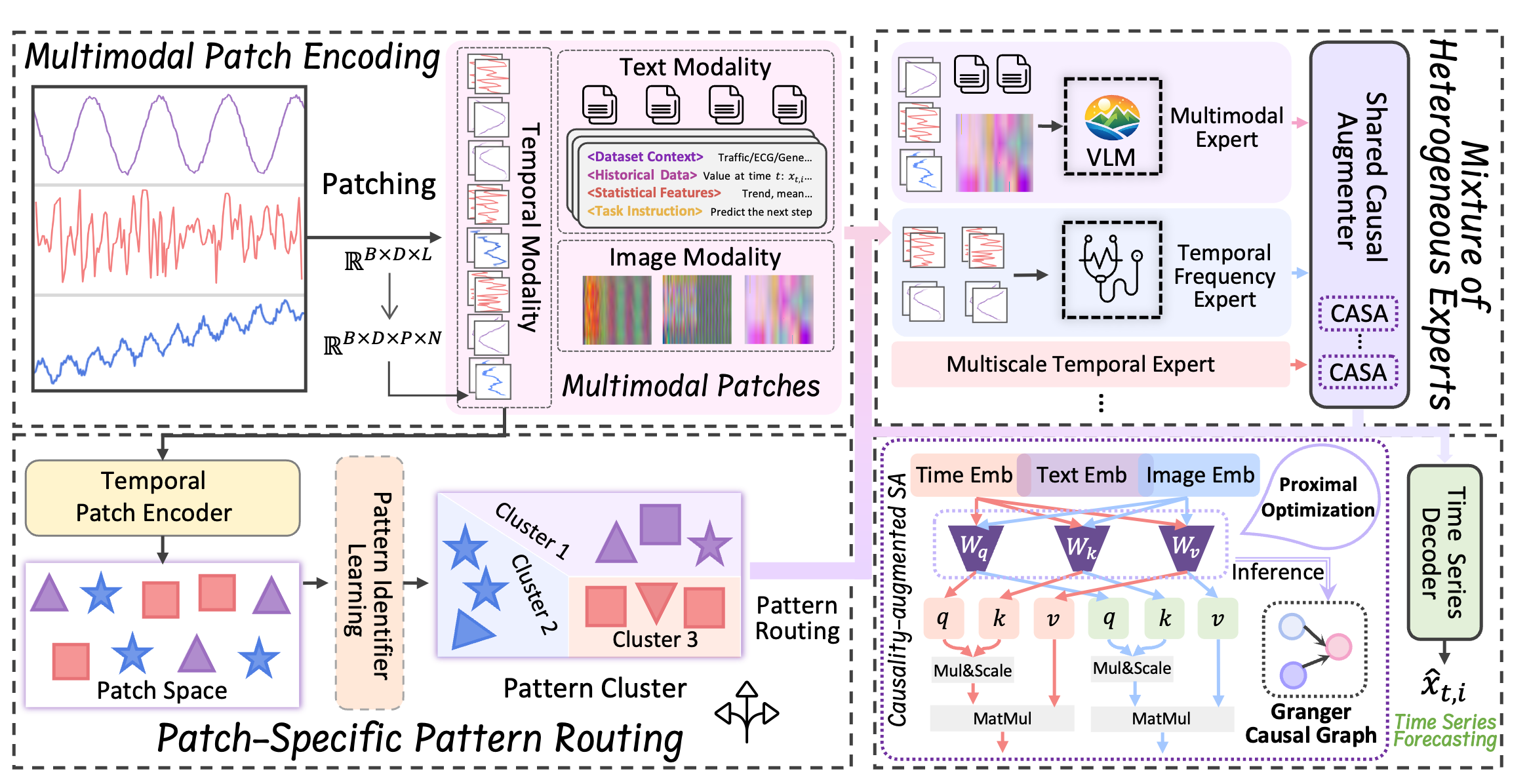}
    \caption{The overall framework of CausalMoE. The architecture unfolds in three modules: (1) Multimodal Patch Encoding transforms time series into aligned temporal, textual, and visual tokens; (2) Patch-Specific Pattern Routing dynamically identifies latent regimes and routes patches to specialized experts; and (3) Mixture of Heterogeneous Experts integrates domain-specific representations via a Causality-Aware Self-Attention mechanism to infer the Granger causal graph.}
    \label{fig:framework}
\end{figure*}

\subsection{Distribution Shifts in Time Series}
These limitations are further amplified by the non-stationarity of real-world time series. Prior work addresses distribution shifts through domain adaptation and generalization \cite{DDN,DUET,DDG-DA,Dish-TS}, adaptive architectures \cite{NST}, and normalization strategies \cite{RevIN}. However, these methods mainly optimize prediction error, which may smooth out regime variations that are informative for Granger causal discovery. CausalMoE instead preserves temporal heterogeneity and uses it as a routing signal to expose regime-specific causal mechanisms to specialized experts.

\section{Methodology}

\subsection{Problem Formulation}
\textbf{Multivariate Time Series}. Let $\mathbf{X} = \{\mathbf{x}_{1:T,1}, \ldots, \mathbf{x}_{1:T,D}\} \in \mathbb{R}^{T \times D}$ denote a multivariate time series instance, with $T$ time steps and $D$ variables. To align with Granger causal discovery, we construct historical segments using a sliding-window strategy with each input segment  $\mathbf{x}_{<t,1:D}=\{\mathbf{x}_{t-L,1:D},\dots,\mathbf{x}_{t-1,1:D}\}$ and to predict the corresponding future time points $\mathbf{x}_{t,1:D}$ in an autoregressive manner, where $L$ represents the length of the look-back window.

\noindent\textbf{Granger Causal Discovery}. Granger causality \cite{GC} is widely used for learning causality from TS data. It is a statistical concept of causality that's based on prediction: time-series $\boldsymbol{x}_j$ Granger causes time-series $i$ when the past values of time-series $\boldsymbol{x}_j$ aid in the prediction of the current and future status of time-series $\boldsymbol{x}_i$.

With advances in multimodal representation learning, time series are no longer restricted to a single temporal modality. To push GCD towards complex real-world applications, we introduce Multimodal Granger Causality (MGC), which aims to infer more interpretable Granger causality by mining joint information across time-series modalities. For a dynamic system represented by $\mathcal{X}$, we extend each univariate segment $\mathbf{x}_{<t,i}$ to incorporate the text modality $\boldsymbol{p}_{t,i}$ and the image modality $\boldsymbol{i}_{t,i}$. We assume each sampled variable $x_{t,i}$ is generated by the following model:
\begin{equation}
    \hat{x}_{t,i}=f_i(\mathbf{x}_{<t,1},\boldsymbol{p}_{t,1},\boldsymbol{i}_{t,1},...,\mathbf{x}_{<t,D},\boldsymbol{p}_{t,D},\boldsymbol{i}_{t,D})+\epsilon_{t,i}
    \label{eq:prediction}
\end{equation}
where $f_i(\cdot)$ is a function mapping the past values and multimodal features of all $D$ time series to variable $i$ and the $\epsilon_{t,i}$ is an independent noise item. 

\begin{definition}
    \textit{\textbf{Multimodal Granger Causal Discovery}. Time series $j$ Granger non-causes $i$ if for all $(\mathbf{x}_{<t,1},...,\mathbf{x}_{<t,D})$, $(\boldsymbol{p}_{t,1},\dots,\boldsymbol{p}_{t,D})$, $(\boldsymbol{i}_{t,1},\dots,\boldsymbol{i}_{t,D})$ and  $\mathbf{x}^\prime_{<t,j}\neq \mathbf{x}_{<t,j}$, $\boldsymbol{p}^\prime_{t,j}\neq \boldsymbol{p}_{t,j}$, $\boldsymbol{i}^\prime_{t,j}\neq \boldsymbol{i}_{t,j}$:}
\begin{equation}
    \begin{aligned}
        &f_i(\mathcal{X}_{t,1},\ldots,\mathbf{x}_{<t,j},\boldsymbol{p}_{t,j},\boldsymbol{i}_{t,j}\ldots,\mathcal{X}_{t,D})\\
=&f_i(\mathcal{X}_{t,1},\ldots,\mathbf{x}^\prime_{<t,j},\boldsymbol{p}^\prime_{t,j},\boldsymbol{i}^\prime_{t,j},\ldots,\mathcal{X}_{t,D})
    \end{aligned}
\end{equation}
\textit{where $\mathcal{X}_{t,i}=(\mathbf{x}_{<t,i},\boldsymbol{p}_{t,i},\boldsymbol{i}_{t,i})$.}
\end{definition}

We emphasize that our multimodal Granger causal discovery remains within the predictive GC framework \cite{NGC}. The textual prompts and visual representations are deterministic views of the observed historical time-series patches, rather than additional exogenous variables; thus, they do not introduce extra confounders, mediators, or colliders beyond the observed series. While Granger causality is not equivalent to true causality, it is justified under standard assumptions such as no unobserved variables and no instantaneous effects \cite{peters2017elements,ACD,CUTS}.

\subsection{Multimodal Patching Encoding}
\textbf{Patching}. The input time series segment $\mathbf{x} \in \mathbb{R}^{L\times D}$ is divided into patches of length $P$, resulting in $N=\lfloor \frac{(L-P)}{S} +2 \rfloor$ tokens, where $S$ denotes the stride, defining the non-overlapping region between consecutive patches. 

\begin{figure}[h]
    \centering
    \includegraphics[width=\linewidth]{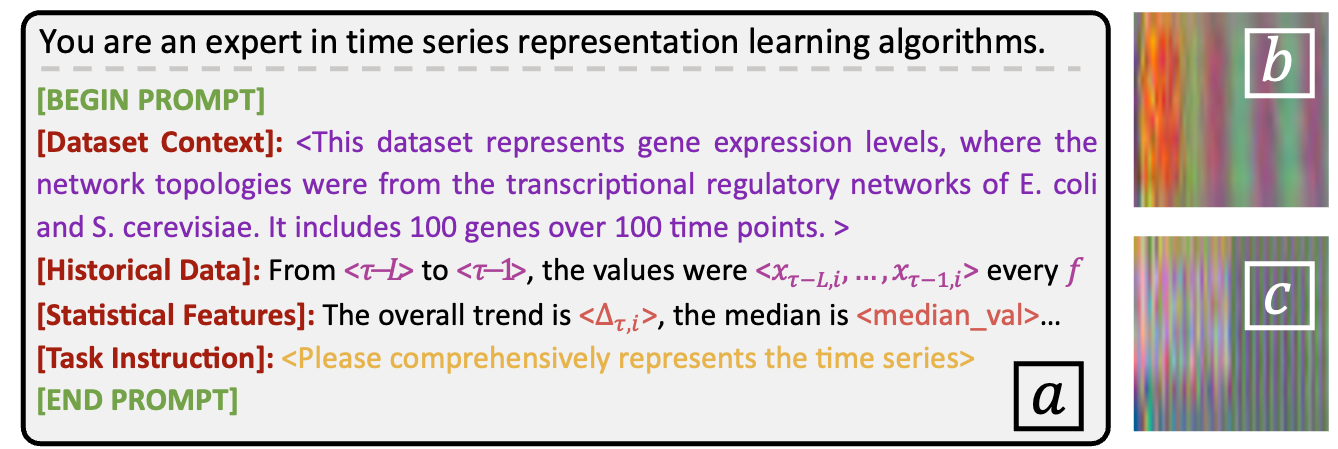}
    \caption{Examples of multimodal patch encoding.
(a) Text prompt template with dynamically instantiated time-series attributes.
(b)--(c) Visual modalities transformed from VAR and Lorenz-96 datasets.}
    \label{fig:prompt}
\end{figure}

\noindent\textbf{Time Series to Prompt}. We wrap the time series patch $\mathbf{p}_i \in \mathbb{R}^{D \times P}$ into prompts $\boldsymbol{\mathcal{P}}_t = \{\boldsymbol{p}_{t,1}, \ldots , \boldsymbol{p}_{t,D}\} \in \mathbb{R}^{L_p\times D}$ along with variables, where $L_p$ denotes the maximal time lag. Each prompt $\boldsymbol{p}_{t,i}$ has $L_p$ elements as shown in Fig.~\ref{fig:prompt}. To avoid benchmark identity leakage, the prompt template is deliberately anonymized. In practice, we identify four key components of effective prompts for LLM/VLM, which are dynamically instantiated with: (1) dataset context to provide domain-specific background, (2) historical data to preserve temporal continuity, (3) statistical features (e.g., trends, medians) to enhance pattern recognition, and (4) task instruction to guide the transformation of patch embeddings.

\noindent\textbf{Time Series to Image}. The patch is resized to the desired image dimensions \( (H, W) \) using bilinear interpolation to construct images \cite{Time-VLM,VisionTS}. For a target pixel \( (x, y) \), the interpolated value \( \mathbf{I}(x, y) \) is computed as follows:

\vspace{-1em}
\begin{equation}
    \mathbf{I}(x, y) = \sum_{i=1}^{2} \sum_{j=1}^{2} \mathbf{I}(x_i, y_j) \cdot w_{ij},
\end{equation}
\vspace{0em}
\begin{equation}
    \mathbf{I}{\text{norm}} = 255 \cdot \frac{\mathbf{I}_{\text{raw}} - \text{Min}(\mathbf{I}_{\text{raw}})}{\text{Max}(\mathbf{I}_{\text{raw}}) - \text{Min}(\mathbf{I}_{\text{raw}}) + \epsilon},
\end{equation}
where \( (x_i, y_j) \) are the coordinates of the four nearest neighbors, \( w_{ij} \) are weights based on relative distances, and \(\epsilon = 10^{-5}\) prevents division by zero. Pixel values are scaled to \([0, 255]\) via min-max normalization, producing the normalized image \(\mathbf{I}_{\text{norm}} \in \mathbb{R}^{B \times C \times H \times W}\) (\( C \) is the number of channels). This ensures alignment with the VLM vision encoder's input distribution for effective feature extraction. 

\subsection{Patch-Specific Pattern Routing}

Following evolving pattern modelling in TS \cite{Pattern-Specific}, we introduce a Patch-Specific Pattern Routing (PSPR) module that dynamically classifies patches based on their distributional characteristics, enabling adaptive modelling under distribution shifts. PSPR leverages subspace clustering to detect concept shifts across multiple subspaces and iteratively refines to improve representation learning and pattern discrimination.

Formally, we construct a set of subspace bases, denoted as $\mathcal{B}=\{\mathcal{B}^1,\mathcal{B}^2,\dots,\mathcal{B}^K\}$, to represent the latent $K$  subspaces, with each $\mathcal{B}^j\in \mathbb{R}^{q\times d}$, $\left\|\mathcal{B}^j_u\right\|_2= 1$ for all $u = 1, \dots d$ and $j=1, \dots, K$. To enforce geometric distinctiveness, we regularize constraints on $\mathcal{B}$:
\begin{equation}
    \mathcal{L}_{Ortho}=\frac{1}{2} \left\| \mathcal{B}^T \mathcal{B} \odot \mathbf{I} - \mathbf{I} \right\|_F^2+\frac{1}{2} \left\| \mathcal{B}^T \mathcal{B} \odot \mathbf{O} \right\|_F^2
\end{equation}
where $\odot$ denotes the Hadamard product, $\mathbf{I}$ is an identity matrix of size $Kd \times Kd$, and $\mathbf{O}$ is a matrix with all off-diagonal d-size blocks set to 1 and diagonal blocks set to 0. The first term of $L_{bases}$ enforces unit norms on the basis vectors of $\mathcal{B}$, while the second term ensures the dissimilarity between different subspaces. The patch router $\mathbf{R}$ is learned to assess the relationship between the embedded representation $\mathbf{z}$ and the subspace bases $\mathcal{B}$. The affinity $r_{ij}$, representing the probability that the embedded $\mathbf{z}_i$ belongs to the $j$-th subspace, is defined as: 
\begin{equation}
    r_{ij} = \frac{\left\| \mathbf{z}_i^T \mathbf{D}^{(j)} \right\|_F^2 + \eta d}{\sum_j (\left\| \mathbf{z}_i^T \mathbf{D}^{(j)} \right\|_F^2 + \eta d)}
\label{eq:patten_routing}
\end{equation}
where $\eta$ is a parameter controlling the smoothness, fixed to the same value as $d$. To emphasize more confident assignments, we refine the subspace affinity $\hat{r}_{ij}$ by weighting high-confidence:
\begin{equation}
\label{eq:subspace refinement}
\hat{r}_{ij} = \frac{r_{ij}^2 / \sum_i r_{ij}}{\sum_j (r_{ij}^2 / \sum_i r_{ij})}.
\end{equation}

By directly analyzing the intrinsic properties of each patch and clustering them into distinct groups based on their latent patterns, CausalMoE leverages discriminative features such as temporal seasonality, trend components, and spectral variations to dynamically route each patch to the most appropriate domain-specific expert tailored for that regime. The objective of PSPR is a combination of regularization on $\mathcal{B}$ and the Kullback-Leibler divergence on $\mathbf{S}$: 

\begin{equation}
    \mathcal{L}_{PR}=\alpha\mathcal{L}_{Ortho}+\beta\sum_i \sum_j \hat{r}_{ij} log\frac{\hat{r}_{ij}}{r_{ij}}
\end{equation}
where $\hat{r}_{ij}$ denotes the predicted assignment, and $r_{ij}$ represents the true distribution of the patch-specific routing.

\subsection{Mixture of Heterogeneous Experts}
To overcome the limitations of Uniform Distribution Modeling (UDM) in handling evolving patterns, we introduce the Pattern-Routing Mixture of Heterogeneous Experts (MoHE). Guided by routing assignments $\mathbf{S}$ from the PSPR module, MoHE applies specialized experts to the feature tensor $z \in \mathbb{R}^{D \times N \times d}$  to enable adaptive causal modeling. It consists of:

\noindent\textbf{Time Series Domain Experts}.
To fully exploit the MoE paradigm, we introduce Heterogeneous Time Series Domain Experts, each incorporating distinct learning mechanisms tailored to specific temporal characteristics. A pattern-specific router dynamically assigns patches to the most suitable expert based on intrinsic features, enabling regime-aware modeling. We outline four such domain-specific experts below.

\textbf{\textit{Semantic Expert}}. The Semantic Expert maps each time-series patch into a language-shaped representation space by wrapping it into a structured prompt (Fig.~\ref{fig:prompt}a) containing statistical summaries and task context, and then encoding it with a frozen LLM. Unlike numerical and visual experts, it captures prompt-level semantic regularities such as trends, noise patterns, and statistical motifs, which provide complementary representation cues for causal discovery under short or ambiguous trajectories:

\begin{equation}
    E_{\text{sem}} = \text{LLM}(\mathbf{p}_i, \boldsymbol{\mathcal{P}}_t)
\end{equation}

\textbf{\textit{Multimodal Expert.}} The Multimodal Expert complements the Semantic Expert by capturing shape-level visual structure from the textual prompt and normalized image rendering of the patch (Eq. 4) using a frozen VLM. It provides perceptual priors, such as peak sharpness, envelope morphology, and regime-transition boundaries, which are difficult to encode with language or scalar statistics alone:
\begin{equation}
    E_{\text{mm}} = \text{VLM}(\mathbf{p}_i, \boldsymbol{\mathcal{P}}_t, \mathbf{I}_{\text{norm}})
\end{equation}

\textbf{\textit{Temporal Frequency Expert.}} Periodic and quasi-periodic dependencies can be diluted by patch-local modeling. We introduce a Temporal Frequency Expert that combines time-domain features with Fourier-domain representations following \cite{filterts}, explicitly capturing stationary spectral patterns and dominant frequencies:
\begin{equation}
    E_{\text{freq}} = \text{Temp-Freq}(\mathbf{x}_p, \mathcal{F}(\mathbf{x}_p))
\end{equation}

\textbf{\textit{Multiscale Temporal Expert.}} The Multiscale Temporal Expert aggregates downsampled representations across temporal resolutions \cite{timemixer}, enabling the model to capture both local variations and global trends. It complements the frequency expert by modeling non-stationary trends and hierarchical seasonal patterns:
\begin{equation}
    E_{\text{ms}} = \sum_{s} \text{DownSample}_s(\mathbf{x}_p) \mathbf{W}_s
\end{equation}

\noindent\textbf{Patch Gating Network}. The patch gating network, denoted as $GaN$, computes the gating weights for all experts based on the cluster assignment $\mathbf{r}$ and activates the top-$k$ experts. To ensure smooth transitions across different pattern regions, we use $K=4$ for four experts and select the top-$k=2$ experts for each input patch. The gating weights are computed as:
\begin{align}
\label{eq:patten_softmax}
GaN(\mathbf{r}) = \text{Softmax} (\text{TopK}(\mathbf{\mathbf{r}}))
\end{align}
Here, the top $k$ logits are selected and normalized using the Softmax function to produce the patch weights.

\noindent\textbf{Expert Aggregation}.The final output $\mathbf{h}$ of the MoHE module is a weighted sum of the outputs from all the selected experts, with the weights provided by the gating network:
\begin{align}
\label{eq:pattern_experts}
\mathbf{h} = \sum_{k=1}^K GaN(\mathbf{r}) E_k(\mathbf{z})
\end{align}

The four experts are functionally specialized rather than redundant.
The Semantic Expert captures global statistical semantics and domain priors, the Multimodal Expert extracts local geometric and shape-level patterns, and the Temporal-Frequency and Multiscale experts model spectral and multi-resolution numerical regularities.
The Patch-Specific Pattern Router dynamically assigns each patch to its top-$k$ experts based on distributional signatures (Eqs.~\ref{eq:patten_routing}--\ref{eq:patten_softmax}), and the gating network fuses their outputs (Eq.~\ref{eq:pattern_experts}).

\subsection{Granger Causal Discovery}
Once optimal prediction is achieved, we infer the causal graph by measuring each source variable's contribution to target prediction.

\noindent\textbf{Causal Augmenter}. To tackle the increased complexity of causal source identification introduced by the multimodal setting, we propose a Causality-Aware Self-Attention (CASA) mechanism. CASA is designed to compute attention across variables rather than across time, preserving variable-wise causal interpretability while avoiding information leakage. Specifically, we transpose the input feature matrix $\mathbf{h} \in \mathbb{R}^{B_{token}\times D \times d}$ into $\mathbf{h}^\top \in \mathbb{R}^{B_{token}\times d \times D}$ so that each column corresponds to a distinct variable. Unlike conventional self-attention employing projection matrices in $\mathbb{R}^{d \times d}$, CASA replaces them with variable-level projections $\omega_q, \omega_k, \omega_v \in \mathbb{R}^{D \times D}$:

\begin{align}
    \mathbf{q}=\mathbf{h^\top}\omega_q,\ \mathbf{k}&=\mathbf{h^\top}\omega_k,\ \mathbf{v}=\mathbf{h^\top}\omega_v \in \mathbb{R}^{B_{token}\times d\times D}\\
    \mathbf{M}&=\mathbf{h^\top}\omega_q (\mathbf{h^\top}\omega_k)^\top \in \mathbb{R}^{B_{token}\times d\times d} \\
    \mathit{CASA}(\mathbf{h})&=\operatorname{Softmax}(\mathbf{M})\mathbf{h^\top}\omega_v \in \mathbb{R}^{B_{token}\times d\times D}
\end{align}

CASA aligns with the Granger causality by computing attention across variables, which explicitly encodes variable-to-variable influence and enables direct causal interpretation. Stacking CASA layers forms a Causal Augmenter that captures higher-order dependencies. Unlike prior methods relying on statistical tests or sparsity, CASA introduces three causality-aware projections—query, key, and value—enhancing interpretability.

Causal augmenter leverages the CASA to enhance causal inference.  We adopt a single CASA block repeated for $1\text{–}l$ iterations to simplify the model, improving interpretability and avoiding the complexity of multiple blocks. Experiments confirm the effectiveness of this approach.

\noindent\textbf{Objective.}
The causality can be represented by an adjacency matrix $\omega_v = \{\omega_v^{:j}\}^N_{j=1}$, where $\omega_v^{:j}\neq0$ denotes series $i$ Granger causes $j$ and otherwise. This approach has been thoroughly investigated and shows strong empirical support in recent years \cite{NGC,han2025root}.

We apply a regularization term on $\omega_q, \omega_k, \omega_v$ to the training loss to promote sparsity in the causal matrix $\omega_v$:
\begin{align}
    &\mathcal{L}=\sum_{t=L+1}^{T}(x_{t,i}-f_i(\boldsymbol{\mathcal{X}}_t))^2+\lambda_1 \mathcal{L}_{Pena}+\lambda_2\mathcal{L}_{PR} \\
    &\mathcal{L}_{Pena}=\sum_{j=1}^N\left(\parallel \omega_q^{:j}\parallel_2+\parallel \omega_k^{:j}\parallel_2+\parallel \omega_v^{:j}\parallel_2\right)
    \label{eq:gc_loss}
\end{align}
where $\boldsymbol{\mathcal{X}}_t=\{\mathcal{X}_{t-L},\dots,\mathcal{X}_{t-1}\}$, $\lambda_1$ is the  regularization trade-off and $\lambda_2$ controls the orthogonality constraint.

\noindent\textbf{Optimizing the Penalized Objective.} We use proximal gradient descent \cite{parikh2014proximal} to optimize the nonconvex objectives of Eq. \ref{eq:gc_loss}, which updates the network weights iteratively starting with $\omega_q,\omega_k,\omega_v$ by
\begin{align}
  \omega_{q,k,v}{(i + 1)} &= \text{prox}_{\gamma}  \left( \omega_{q,k,v}{(i)} - \gamma \nabla \mathcal{L}_{pred} (\omega_{q,k,v}{(i)}) \right)  
\end{align}
where $\text{prox}_\gamma$ denotes the proximal operator with step size $\gamma$; $\mathcal{L}_{pred}=\sum_{t=L+1}^{T}(x_{t,i}-f_i(\boldsymbol{\mathcal{X}}_t))^2)$ denotes the convex part of the neural network prediction loss. The proximal step for the input weights involves a group soft-thresholding operation \cite{parikh2014proximal}:
\begin{align}
\text{prox}_{\gamma \rho }(\omega^{:j})=\text{soft} (\omega^{:j}, \gamma \rho) =  (1 - \frac{\rho \gamma}{||\omega^{:j}||_{2}} )_{+} \omega^{:j}
\end{align}
where $(x)_+ = \max(0, x)$. 








\noindent\textbf{Causal Graph Construction}.  
Let \( G = (V, E) \) represent the Granger causal graph, where \( V \) contains \( N \) dependent time series \(( \mathbf{x}_1, \dots, \mathbf{x}_N) \), and \( E \) represents causal relationships. For each \( \mathbf{x}_i \), a prediction function \( f_{i} \) learns to predict \( \mathbf{x}_i \), generating a causal weight matrix \( \omega_{v,i} \in \mathbb{R}^{N \times N} \). The columns of \( \omega_{v,i} \), denoted \( \omega_{v,i}^{:j} \), quantify the causal influence of \( \mathbf{x}_j \) on \( \mathbf{x}_i \). An edge \( \mathbf{x}_j \rightarrow \mathbf{x}_i \) exists if \( \omega_{v,i}^{:j}   \neq 0 \), where:  (1) \( i \neq j \): Past values of \( \mathbf{x}_i \) predict \( \mathbf{x}_j \); (2) \( i = j \): Self-causality, meaning \( \mathbf{x}_i \) predicts its own future.

\section{Experiments}

\subsection{Experimental Setup}
\noindent\textbf{Datasets}. We evaluate the proposed CausalMoE framework on five benchmark datasets:

\textbf{VAR (linear)}. Linear VAR datasets are simulated following:
\begin{equation}
x^t=\sum_{\tau=1}^{\tau_{max}}\mathbf{A}_\tau x^{t-\tau}+e_t,
\end{equation}
where $\mathbf{A}_\tau$ is the sparse coefficients for time lag $\tau$.

\textbf{Lorenz-96 (non-linear)}. As a nonlinear model to simulate climate dynamics, the p-dimensional Lorenz-96 \cite{lorenz} model is defined:
\begin{equation}
    \frac{dx^n_i}{dt}=(\mathbf{x}^{i+1}-\mathbf{x}^{i-2})\mathbf{x}^{i-1}-\mathbf{x}^i+F
\end{equation}
where $x_{-1} = x_{p-1}$, $ x^0 = x_p$ and $x_{p+1} = x_1$. $F$ is a forcing constant controlling the system's chaotic degree.

\textbf{fMRI}. fMRI \cite{fMRI} serves as a benchmark for causal discovery, with realistic simulations of blood-oxygen-level-dependent time series generated through a dynamic causal modeling framework for functional magnetic resonance imaging.

\textbf{DREAM-3 and DREAM-4}. The DREAM-3 and DREAM-4 ~\cite{Dream-4} are publicly available, realistic gene expression datasets, aiming to reconstruct gene regulatory networks from gene expression data. The DREAM-3 includes two Ecoli and three Yeast datasets, and the DREAM-4 consists of five independent gene regulatory datasets.

\noindent\textbf{Baselines}. We perform comparative experiments with seven competitive methods: (1) \textbf{GC} \cite{GC}, a linear model for Granger causality tests. (2) \textbf{PCMCI} \cite{pcmci}, which uses conditional independence tests with optimized conditioning sets. (3) \textbf{NGC} \cite{NGC}, a component-wise LSTM with sparse input weights for non-linear Granger causality. (4) \textbf{CR-VAE} \cite{CR-VAE}, specialized RNNs for Granger causal structure identification. (5) \textbf{CUTS} \cite{CUTS}, a neural GCD method for imputed and high-dimensional data. (6) \textbf{JRNGC} \cite{JRNGC}, which jointly infers causality using a shared encoder-decoder structure. (7) \textbf{KANGCI} \cite{KAN}, which adapts Kolmogorov–Arnold Networks (KANs) to learn causal mechanisms from time series.

\noindent\textbf{Evaluation Metrics}. We adopt four standard evaluation metrics: (1) AUROC, measuring the area under the ROC curve; (2) AUPRC, capturing the area under the precision-recall curve; (3) F1 Score, the harmonic mean of precision and recall and (4) SHD, Structural Hamming Distance, differences between predicted and ground truth.

\noindent\textbf{Evaluation Setting}. We evaluate CausalMoE under a full data supervised setting and a few-shot causal discovery setting, where only a limited number of observations are available for training causal models. This evaluation protocol is designed to assess whether CausalMoE can leverage its pretrained knowledge and multimodal priors to infer reliable causal structures under severe data scarcity.

\begin{table*}[htbp]
\scriptsize
\caption{Overall performance (mean±std.) on synthetic VAR and Lorenz-96 datasets for Granger causal discovery.}
\label{tab:var_lorenz}
\centering
\resizebox{\textwidth}{!}{
  \begin{tabular}{c|c|c|c|c|c|c|c|c|c}
    \toprule
    \textbf{Synthetic Dataset} & \textbf{Metrics} & \textbf{GC} & \textbf{PCMCI} & \textbf{NGC} & \textbf{CR-VAE} & \textbf{CUTS} & \textbf{KANGCI} & \textbf{JRNGC} & \textbf{CausalMoE} \\
    \midrule
    \multirow{4}{*}{\textbf{VAR(20,1000,5)}}  & AUROC ($\uparrow$)  & \numstd{0.602}{0.019} & \numstd{0.660}{0.020} & \numstd{0.753}{0.020} & \numstd{0.929}{0.015} & \numstd{0.944}{0.009} & \numstd{0.828}{0.026} & \textcolor{blue}{\underline{\numstd{0.972}{0.017}}} & \textcolor{red}{\textbf{\numstd{0.989}{0.032}}} \\
     & AUPRC ($\uparrow$)  & \numstd{0.607}{0.013} & \numstd{0.745}{0.016} & \numstd{0.742}{0.016} & \numstd{0.962}{0.020} & \textcolor{blue}{\underline{\numstd{0.985}{0.034}}} & \numstd{0.844}{0.017} & \numstd{0.965}{0.020} & \textcolor{red}{\textbf{\numstd{0.993}{0.018}}} \\
     & F1 ($\uparrow$)  & \numstd{0.605}{0.030} & \numstd{0.696}{0.030} & \numstd{0.735}{0.017} & \numstd{0.924}{0.026} & \numstd{0.966}{0.017} & \numstd{0.858}{0.016} & \textcolor{blue}{\underline{\numstd{0.972}{0.016}}} & \textcolor{red}{\textbf{\numstd{0.986}{0.018}}} \\
     & SHD ($\downarrow$)  & \numstd{40}{4} & \numstd{33}{2} & \numstd{13}{4} & \numstd{8}{2} & \textcolor{blue}{\underline{\numstd{5}{2}}} & \numstd{13}{3} & \numstd{13}{1} & \textcolor{red}{\textbf{\numstd{4}{1}}} \\
    \midrule
    \multirow{4}{*}{\textbf{VAR(20,500,20)}}  & AUROC ($\uparrow$)  & \numstd{0.572}{0.019} & \numstd{0.596}{0.030} & \numstd{0.692}{0.017} & \numstd{0.824}{0.014} & \numstd{0.844}{0.027} & \numstd{0.826}{0.021} & \textcolor{blue}{\underline{\numstd{0.933}{0.015}}} & \textcolor{red}{\textbf{\numstd{0.976}{0.015}}} \\
     & AUPRC ($\uparrow$)  & \numstd{0.593}{0.023} & \numstd{0.591}{0.028} & \numstd{0.671}{0.024} & \numstd{0.833}{0.017} & \numstd{0.842}{0.028} & \numstd{0.807}{0.020} & \textcolor{blue}{\underline{\numstd{0.923}{0.015}}} & \textcolor{red}{\textbf{\numstd{0.977}{0.004}}} \\
     & F1 ($\uparrow$)  & \numstd{0.706}{0.019} & \numstd{0.604}{0.044} & \numstd{0.676}{0.017} & \numstd{0.814}{0.026} & \numstd{0.818}{0.014} & \numstd{0.821}{0.007} & \textcolor{blue}{\underline{\numstd{0.944}{0.012}}} & \textcolor{red}{\textbf{\numstd{0.963}{0.003}}} \\
     & SHD ($\downarrow$)  & \numstd{178}{5} & \numstd{166}{12} & \numstd{97}{6} & \numstd{97}{6} & \textcolor{blue}{\underline{\numstd{21}{4}}} & \numstd{67}{16} & \numstd{59}{8} & \textcolor{red}{\textbf{\numstd{16}{3}}} \\
    \midrule
    \multirow{4}{*}{\textbf{VAR(40,1000,20)}}  & AUROC ($\uparrow$)  & \numstd{0.593}{0.030} & \numstd{0.572}{0.026} & \numstd{0.651}{0.013} & \numstd{0.783}{0.027} & \numstd{0.832}{0.018} & \numstd{0.785}{0.014} & \textcolor{blue}{\underline{\numstd{0.922}{0.005}}} & \textcolor{red}{\textbf{\numstd{0.953}{0.006}}} \\
     & AUPRC ($\uparrow$)  & \numstd{0.575}{0.021} & \numstd{0.586}{0.024} & \numstd{0.639}{0.017} & \numstd{0.848}{0.017} & \numstd{0.843}{0.016} & \numstd{0.811}{0.022} & \textcolor{blue}{\underline{\numstd{0.906}{0.017}}} & \textcolor{red}{\textbf{\numstd{0.962}{0.013}}} \\
     & F1 ($\uparrow$)  & \numstd{0.706}{0.012} & \numstd{0.575}{0.033} & \numstd{0.632}{0.022} & \numstd{0.802}{0.013} & \numstd{0.812}{0.031} & \numstd{0.786}{0.012} & \textcolor{blue}{\underline{\numstd{0.934}{0.011}}} & \textcolor{red}{\textbf{\numstd{0.956}{0.018}}} \\
     & SHD ($\downarrow$)  & \numstd{163}{4} & \numstd{157}{11} & \numstd{84}{1} & \numstd{81}{11} & \numstd{80}{6} & \numstd{60}{10} & \textcolor{blue}{\underline{\numstd{33}{6}}} & \textcolor{red}{\textbf{\numstd{8}{1}}} \\
    \midrule
    \multirow{4}{*}{\textbf{Lorenz(20,1000,10)}}  & AUROC ($\uparrow$)  & \numstd{0.636}{0.027} & \numstd{0.605}{0.022} & \numstd{0.710}{0.018} & \numstd{0.921}{0.013} & \numstd{0.854}{0.018} & \numstd{0.867}{0.020} & \textcolor{blue}{\underline{\numstd{0.936}{0.019}}} & \textcolor{red}{\textbf{\numstd{0.986}{0.010}}} \\
     & AUPRC ($\uparrow$)  & \numstd{0.606}{0.014} & \numstd{0.630}{0.017} & \numstd{0.711}{0.026} & \numstd{0.896}{0.020} & \numstd{0.865}{0.023} & \numstd{0.878}{0.008} & \textcolor{blue}{\underline{\numstd{0.951}{0.017}}} & \textcolor{red}{\textbf{\numstd{0.988}{0.017}}} \\
     & F1 ($\uparrow$)  & \numstd{0.603}{0.026} & \numstd{0.640}{0.020} & \numstd{0.725}{0.023} & \numstd{0.899}{0.005} & \numstd{0.819}{0.017} & \numstd{0.869}{0.023} & \textcolor{blue}{\underline{\numstd{0.928}{0.009}}} & \textcolor{red}{\textbf{\numstd{0.985}{0.017}}} \\
     & SHD ($\downarrow$)  & \numstd{47}{1} & \numstd{42}{2} & \numstd{30}{4} & \textcolor{blue}{\underline{\numstd{12}{1}}} & \numstd{15}{4} & \numstd{12}{2} & \numstd{12}{3} & \textcolor{red}{\textbf{\numstd{6}{1}}} \\
    \midrule
    \multirow{4}{*}{\textbf{Lorenz(20,500,20)}}  & AUROC ($\uparrow$)  & \numstd{0.544}{0.019} & \numstd{0.572}{0.015} & \numstd{0.653}{0.023} & \numstd{0.857}{0.018} & \numstd{0.815}{0.038} & \numstd{0.777}{0.018} & \textcolor{blue}{\underline{\numstd{0.899}{0.020}}} & \textcolor{red}{\textbf{\numstd{0.950}{0.008}}} \\
     & AUPRC ($\uparrow$)  & \numstd{0.573}{0.012} & \numstd{0.582}{0.008} & \numstd{0.660}{0.011} & \numstd{0.869}{0.018} & \numstd{0.866}{0.016} & \numstd{0.786}{0.014} & \textcolor{blue}{\underline{\numstd{0.929}{0.022}}} & \textcolor{red}{\textbf{\numstd{0.954}{0.016}}} \\
     & F1 ($\uparrow$)  & \numstd{0.692}{0.018} & \numstd{0.574}{0.016} & \numstd{0.730}{0.011} & \numstd{0.569}{0.019} & \numstd{0.815}{0.026} & \numstd{0.765}{0.012} & \textcolor{blue}{\underline{\numstd{0.920}{0.004}}} & \textcolor{red}{\textbf{\numstd{0.950}{0.006}}} \\
     & SHD ($\downarrow$)  & \numstd{196}{2} & \numstd{184}{11} & \numstd{139}{5} & \numstd{104}{8} & \textcolor{blue}{\underline{\numstd{68}{18}}} & \numstd{82}{7} & \numstd{78}{9} & \textcolor{red}{\textbf{\numstd{20}{3}}} \\
    \midrule
    \multirow{4}{*}{\textbf{Lorenz(40,1000,20)}}  & AUROC ($\uparrow$)  & \numstd{0.557}{0.020} & \numstd{0.553}{0.014} & \numstd{0.710}{0.017} & \numstd{0.739}{0.019} & \numstd{0.828}{0.006} & \numstd{0.722}{0.017} & \textcolor{blue}{\underline{\numstd{0.913}{0.009}}} & \textcolor{red}{\textbf{\numstd{0.939}{0.004}}} \\
     & AUPRC ($\uparrow$)  & \numstd{0.554}{0.009} & \numstd{0.548}{0.031} & \numstd{0.693}{0.018} & \numstd{0.806}{0.016} & \numstd{0.824}{0.006} & \numstd{0.762}{0.015} & \textcolor{blue}{\underline{\numstd{0.905}{0.015}}} & \textcolor{red}{\textbf{\numstd{0.947}{0.016}}} \\
     & F1 ($\uparrow$)  & \numstd{0.568}{0.015} & \numstd{0.573}{0.042} & \numstd{0.757}{0.012} & \numstd{0.762}{0.025} & \numstd{0.772}{0.015} & \numstd{0.761}{0.029} & \textcolor{blue}{\underline{\numstd{0.909}{0.019}}} & \textcolor{red}{\textbf{\numstd{0.942}{0.011}}} \\
     & SHD ($\downarrow$)  & \numstd{168}{7} & \numstd{171}{11} & \numstd{91}{8} & \numstd{93}{9} & \numstd{78}{10} & \numstd{152}{10} & \textcolor{blue}{\underline{\numstd{31}{10}}} & \textcolor{red}{\textbf{\numstd{25}{4}}} \\
    \bottomrule
  \end{tabular}
  }
\end{table*}

\subsection{Overall Causal Discovery Performance}

\begin{figure}[ht]
    \centering
    \includegraphics[width=0.8\linewidth]{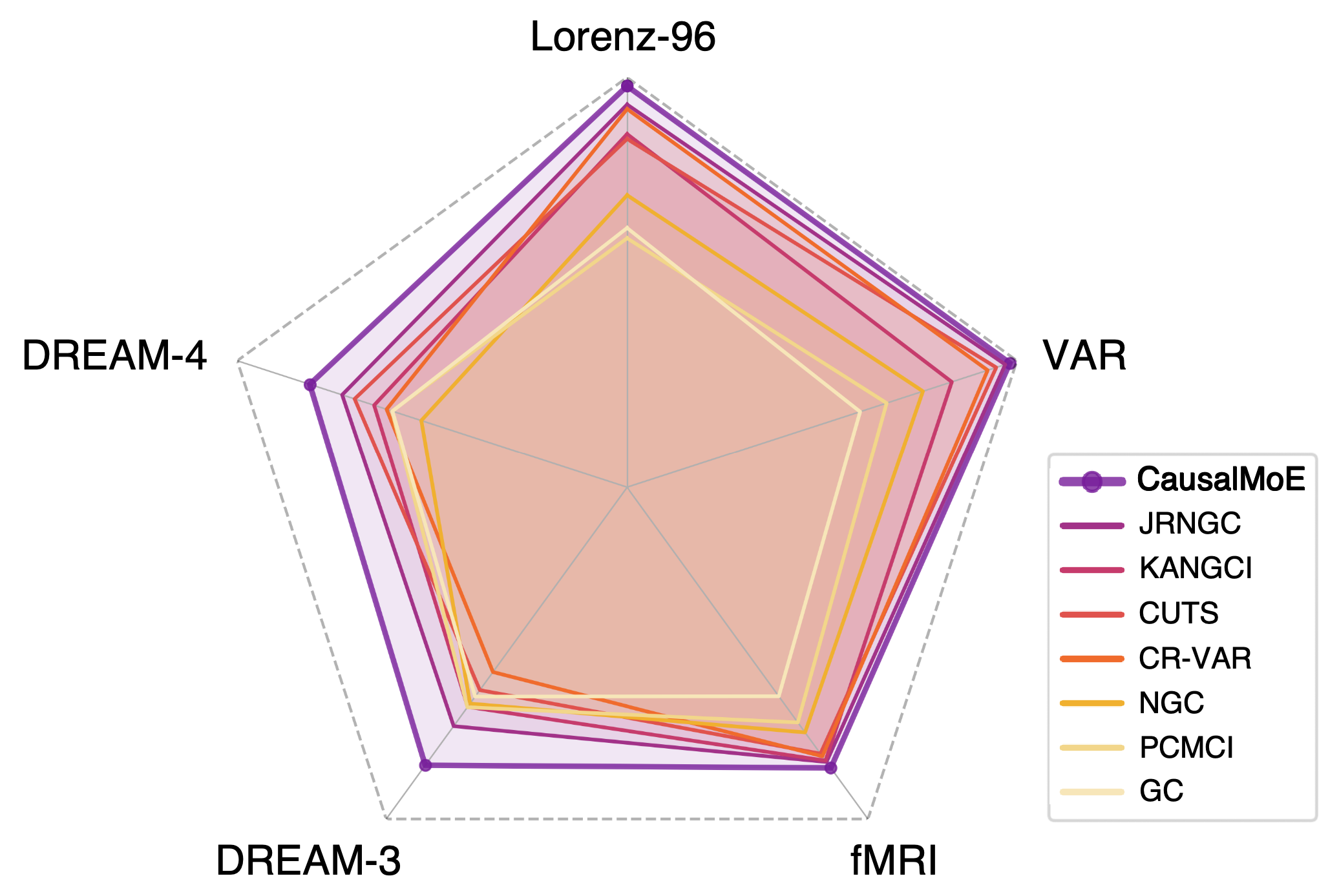}
    \caption{Overall performance of CausalMoE on five benchmarks compared to baselines.}
    \label{fig:overall_performance}
\end{figure}

We visualize the performance of CausalMoE across all five benchmark datasets, as illustrated in Fig.~\ref{fig:overall_performance}. CausalMoE consistently achieves superior performance across all five benchmark datasets. It significantly outperforms traditional Granger causal discovery methods and demonstrates competitive advantages over recent learning-based approaches such as CUTS and JRNGC. Notably, the improvement is particularly evident on complex datasets such as DREAM-3 and DREAM-4, highlighting the benefit of integrating LLM-driven semantic signals with time series dynamics.

\begin{figure*}[ht]
    \centering
    \includegraphics[width=\linewidth]{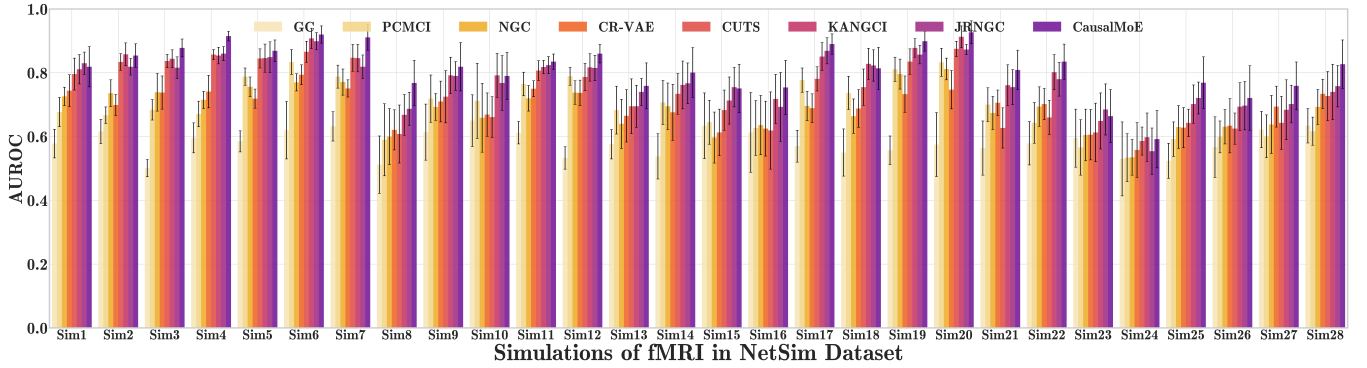}
    \caption{AUROC performance on the fMRI NetSim benchmark across 28 simulated brain connectivity settings.
}
    \label{fig:fMRI}
\end{figure*}

\subsection{Experiments on Synthetic Benchmarks}

\noindent\textbf{VAR}. We simulated $N\in\{20,40\}$ time series over $T \in \{500,1000\}$ observations with a maximum time lag of $\tau \in \{5,20\}$. As shown in Table~\ref{tab:var_lorenz}, CausalMoE consistently achieves the highest AUROC, AUPRC, and F1 scores, along with the lowest SHD across all VAR settings. Even in the most challenging case ($N=40$, $\tau=20$), CausalMoE surpasses all baselines by a notable margin. Moreover, it shows more stable SHD than JRNGC, showing its robustness and reliability in extracting accurate structures under high-dimensional, long-range dependencies.

\noindent\textbf{Lorenz-96}. The Lorenz-96 system is used to evaluate model robustness under nonlinear chaotic dynamics. We vary $N \in \{20, 40\}$ and set the forcing constant $F \in \{10, 20\}$, where higher $F$ introduces stronger chaos. CausalMoE consistently outperforms all baselines across all the metrics. Particularly under chaotic regimes (e.g., $F=20$), traditional and neural methods degrade notably, while CausalMoE maintains strong structure recovery. This superior performance can be attributed to our CASA, which enables variable-wise causal attention and facilitates interpretable structure learning.

\subsection{Experiments on Real-world Benchmarks}
\noindent\textbf{fMRI}. We evaluate CausalMoE on the simulated fMRI BOLD dataset, which contains 28 simulations. Each simulation includes time series data from 50 subjects, covering diverse brain connectivity patterns. Unlike previous studies that focused on a limited subset of simulations, we conduct a comprehensive evaluation across all settings. As shown in Fig.~\ref{fig:fMRI}, CausalMoE achieves competitive AUROC scores in most simulations, and performs favorably in 22 out of 28 cases. While other methods, such as JRNGC, also show strong results in specific conditions, CausalMoE offers consistent performance with lower variance, suggesting better adaptability across a range of causal patterns. This may be attributed to the incorporation of world knowledge through LLM-based representations, which can help distinguish subtle causal relationships in complex scenarios.

\begin{table}[ht]
  \centering
  \caption{AUROC for the sub-datasets in DREAM-3 and in DREAM-4.}
  \label{tab:DREAM}
  \vspace{-4pt}
\resizebox{0.48\textwidth}{!}{
  \begin{tabular}{lccccc}
    \toprule
    \multirow{2}{*}{\textbf{Models}} &   \multicolumn{5}{c}{\textbf{DREAM-3}}\\
    \cmidrule{2-6} 
& {\textbf{Ecoli-1}} & {\textbf{Ecoli-2}} & {\textbf{Yeast-1}} & {\textbf{Yeast-2}} & {\textbf{Yeast-3}} \\
    \midrule
      GC & \numstd{0.556}{0.012} & \numstd{0.635}{0.024} & \numstd{0.646}{0.006} & \numstd{0.637}{0.021} & \numstd{0.560}{0.003} \\
      PCMCI & \numstd{0.615}{0.023} & \numstd{0.628}{0.027} & \numstd{0.625}{0.028} & \numstd{0.623}{0.011} & \numstd{0.627}{0.009} \\
      NGC & \numstd{0.645}{0.031} & \numstd{0.635}{0.036} & \numstd{0.594}{0.014} & \numstd{0.596}{0.030} & \numstd{0.597}{0.005} \\
      CR-VAE & \numstd{0.655}{0.039} & \numstd{0.627}{0.004} & \numstd{0.633}{0.015} & \numstd{0.591}{0.039} & \numstd{0.600}{0.021} \\
      CUTS & \numstd{0.641}{0.013} & \numstd{0.562}{0.021} & \numstd{0.584}{0.006} & \numstd{0.503}{0.018} & \numstd{0.535}{0.041} \\
      KANGCI & \numstd{0.662}{0.018} & \numstd{0.622}{0.014} & \textcolor{blue}{\underline{\numstd{0.694}{0.005}}} & \numstd{0.665}{0.035} & \numstd{0.617}{0.024} \\
      JRNGC & \textcolor{blue}{\underline{\numstd{0.713}{0.006}}} & \textcolor{blue}{\underline{\numstd{0.673}{0.013}}} & \numstd{0.634}{0.031} & \textcolor{blue}{\underline{\numstd{0.694}{0.021}}} & \textcolor{blue}{\underline{\numstd{0.684}{0.032}}} \\
      \textbf{CausalMoE} & \textcolor{red}{\textbf{\numstd{0.845}{0.024}}} & \textcolor{red}{\textbf{\numstd{0.795}{0.032}}} & \textcolor{red}{\textbf{\numstd{0.774}{0.009}}} & \textcolor{red}{\textbf{\numstd{0.750}{0.037}}} & \textcolor{red}{\textbf{\numstd{0.773}{0.026}}} \\
  
    \toprule
    \multirow{2}{*}{\textbf{Models}} &   \multicolumn{5}{c}{\textbf{DREAM-4}}\\
    \cmidrule{2-6} 
& {\textbf{Gene-1}} & {\textbf{Gene-2}} & {\textbf{Gene-3}} & {\textbf{Gene-4}} & {\textbf{Gene-5}} \\
    \midrule
      GC & \numstd{0.606}{0.011} & \numstd{0.558}{0.024} & \numstd{0.501}{0.008} & \numstd{0.495}{0.018} & \numstd{0.505}{0.036} \\
      PCMCI & \numstd{0.605}{0.029} & \numstd{0.552}{0.034} & \numstd{0.514}{0.015} & \numstd{0.515}{0.027} & \numstd{0.505}{0.005} \\
      NGC & \numstd{0.526}{0.030} & \numstd{0.577}{0.008} & \numstd{0.477}{0.035} & \numstd{0.540}{0.014} & \numstd{0.548}{0.013} \\
      CR-VAE & \numstd{0.607}{0.019} & \numstd{0.533}{0.018} & \numstd{0.540}{0.024} & \numstd{0.534}{0.002} & \numstd{0.563}{0.043} \\
      CUTS & \numstd{0.693}{0.014} & \numstd{0.644}{0.016} & \textcolor{blue}{\underline{\numstd{0.654}{0.018}}} & \numstd{0.633}{0.022} & \textcolor{blue}{\underline{\numstd{0.665}{0.006}}} \\
      KANGCI & \numstd{0.641}{0.026} & \numstd{0.624}{0.026} & \numstd{0.639}{0.033} & \numstd{0.646}{0.016} & \numstd{0.628}{0.009} \\
      JRNGC & \textcolor{blue}{\underline{\numstd{0.739}{0.019}}} & \textcolor{blue}{\underline{\numstd{0.733}{0.013}}} & \numstd{0.647}{0.013} & \textcolor{blue}{\underline{\numstd{0.647}{0.023}}} & \numstd{0.648}{0.003} \\
      \textbf{CausalMoE} & \textcolor{red}{\textbf{\numstd{0.829}{0.014}}} & \textcolor{red}{\textbf{\numstd{0.801}{0.012}}} & \textcolor{red}{\textbf{\numstd{0.719}{0.016}}} & \textcolor{red}{\textbf{\numstd{0.753}{0.025}}} & \textcolor{red}{\textbf{\numstd{0.758}{0.019}}} \\
  
    \bottomrule
  \end{tabular}
}
\vspace{-10pt}
\end{table}

\noindent\textbf{DREAM-3, DREAM-4}.
We evaluate the performance of CausalMoE on two widely used benchmark datasets for causal discovery from gene expression data: DREAM-3 and DREAM-4 in silico challenges. Each dataset contains five sub-datasets with ground-truth Granger causal graphs. The evaluation metric is AUROC. As shown in Table~\ref{tab:DREAM}, CausalMoE achieves the highest AUROC scores in all five sub-datasets. Compared to baselines, CausalMoE demonstrates consistently better performance across both bacterial and yeast systems. Table~\ref{tab:DREAM} also reports results on the DREAM-4 dataset, where CausalMoE shows leading performance in all five sub-datasets. In contrast, the second-best method JRNGC obtains lower scores in all cases, and classical methods show performance degradation, particularly under the limited observation setting of DREAM-4. These results suggest that CausalMoE is competitive across both datasets with complex structures and limited time points.

\begin{figure*}
    \centering
    \includegraphics[width=\linewidth]{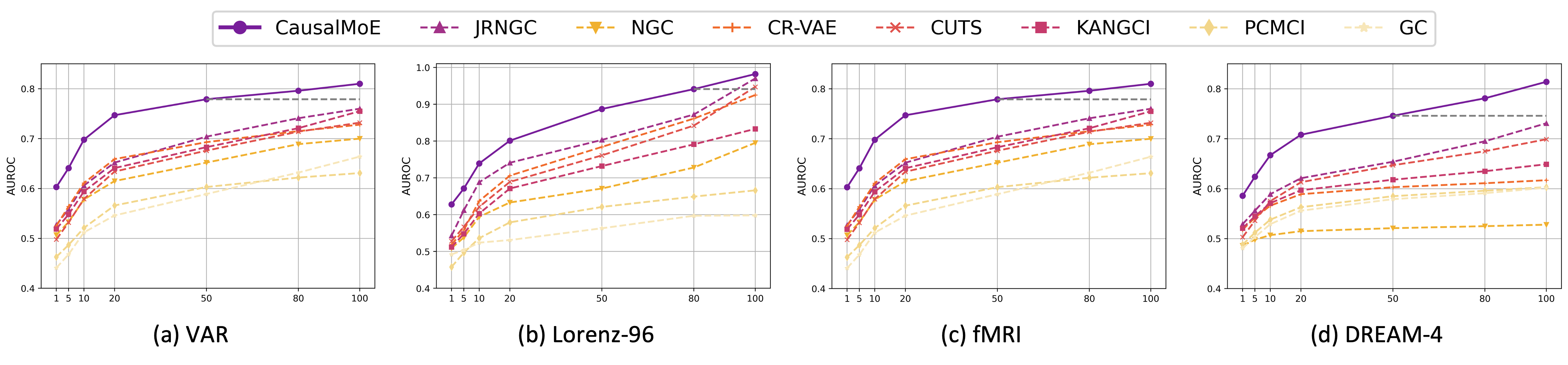}
    \caption{AUROC performance vs. training data ratio (\%) across four benchmarks. CausalMoE (purple line) exhibits strong few-shot inference capabilities, matching the performance of baselines trained on full data with only a fraction of observations.}
    \label{fig:few_shot}
\end{figure*}

\subsection{Few-shot Causal Discovery}


A central motivation of CausalMoE is reliable causal discovery under severe data scarcity, which is common in scientific domains where long and clean observations are difficult to obtain. 
We evaluate data efficiency by varying the training ratio from $1\%$ to $100\%$ across four benchmarks and report AUROC. 
As shown in Fig.~\ref{fig:few_shot}, CausalMoE consistently outperforms all baselines, with the largest advantage in the extreme few-shot regime. 
On the challenging DREAM-4 dataset, CausalMoE achieves an AUROC above $0.6$ using only $5\%$ of the training data, matching competitive baselines such as CUTS and KANGCI trained with $50\%$--$100\%$ data. 
Similar trends are observed on VAR and Lorenz-96, where CausalMoE reaches stable and near-converged performance with substantially fewer observations. 
In contrast, conventional neural methods degrade sharply below $10\%$ training data, indicating that purely numerical supervision is prone to overfitting noisy short sequences and recovering unstable causal structures. 
These results demonstrate that CausalMoE is particularly effective when observations are scarce, rather than only improving performance in data-rich settings.

This data efficiency comes from two complementary factors. 
First, frozen LLM and VLM experts provide semantic and visual priors that regularize causal estimation when numerical observations are sparse. 
Second, Patch-Specific Pattern Routing disentangles heterogeneous temporal regimes before causal estimation, allowing each expert to model a more stable sub-distribution. 
Together, these mechanisms help CausalMoE recover robust causal structures in low-resource settings where purely numerical methods are prone to noisy or spurious links.

\begin{figure}[h]
    \centering
    \includegraphics[width=\linewidth]{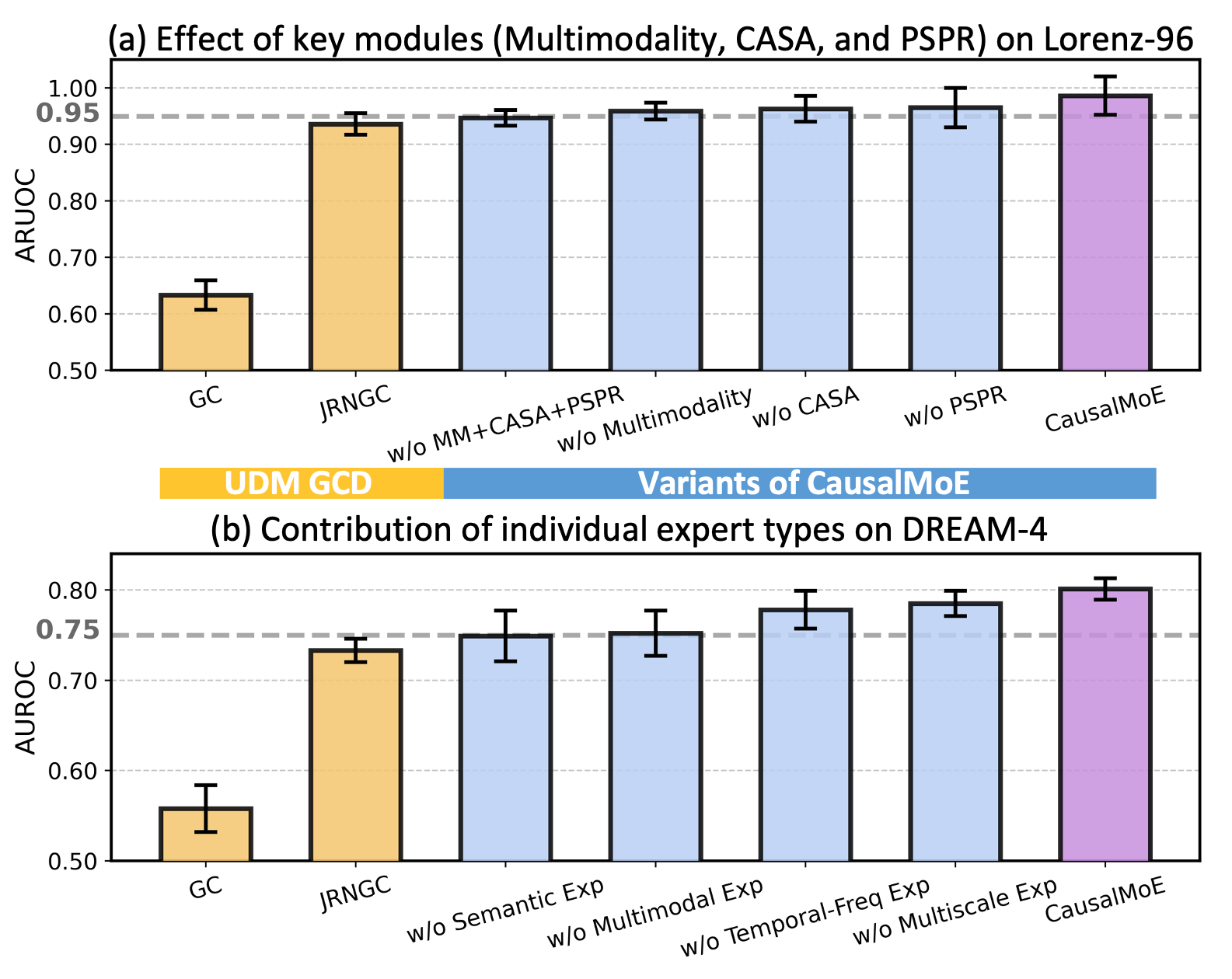}
    \caption{Ablation analysis of CausalMoE. 
(a) Effects of core modules on Lorenz-96. 
(b) Contributions of heterogeneous experts on DREAM-4.}
    \label{fig:ablation_modules}
    \vspace{-10pt}
\end{figure}

\begin{figure*}
    \centering
    \includegraphics[width=\linewidth]{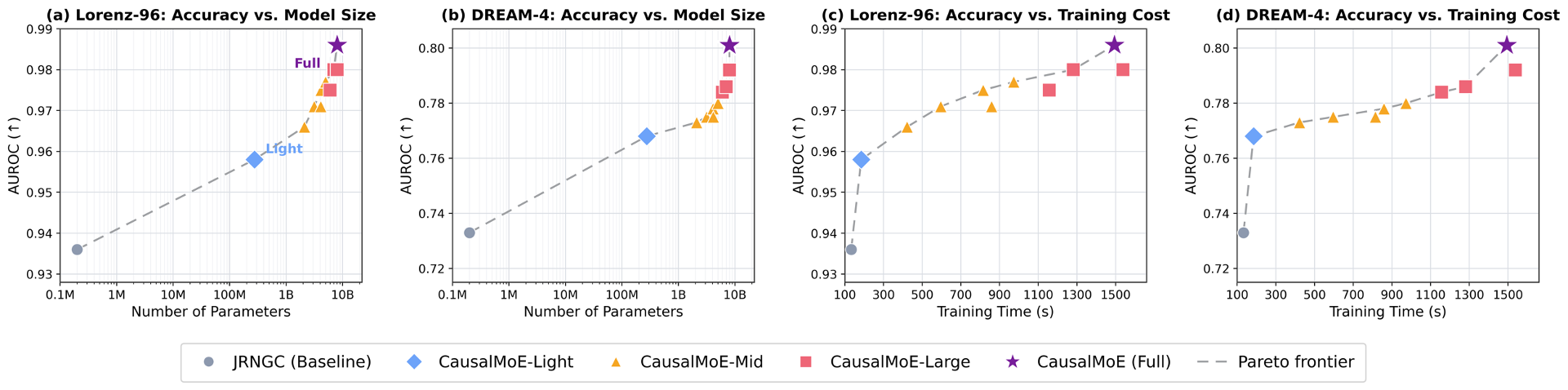}
    \caption{Efficiency--accuracy trade-off of CausalMoE across foundation-model backbones of varying scale, evaluated against parameter count (a, b) and
training cost (c, d) on Lorenz-96 and DREAM-4 with the Pareto frontier traced as a dashed curve.
}
    \label{fig:pareto_frontier}
\end{figure*}

\subsection{Ablation Study}

To isolate the contribution of each design, we conduct comprehensive
ablations on a representative synthetic benchmark (Lorenz-96) and a real-world
benchmark (DREAM-4). We study three aspects: (i) the necessity of the three
core modules, (ii) the contribution and complementarity of the four
heterogeneous experts, and (iii) the efficiency--accuracy trade-off induced by
the model scale.

\noindent\textbf{Are the core components necessary for Granger causal discovery?}
We examine the three major components of CausalMoE: the pattern router,
multimodal transformation, and causality-aware self-attention (CASA). As shown
in Fig.~\ref{fig:ablation_modules}, removing any of them consistently degrades
both causal discovery and downstream forecasting. The \textit{w/o Routing}
variant replaces dynamic assignment with a uniform expert mixture, yielding
lower AUROC and higher forecasting error---confirming that routing is a
necessary mechanism for disentangling heterogeneous regimes before causal
estimation, not merely an architectural add-on. The \textit{w/o Multimodality}
variant relies only on raw numerical inputs, and its drop shows that textual
and visual views provide complementary priors that help regularize dependency
estimation, especially when numerical correlations alone are ambiguous or noisy.
Finally, the degradation of \textit{w/o CASA} indicates that expert
representations alone are insufficient: CASA is needed to explicitly capture
sparse dependencies and translate routed multimodal features into
an interpretable causal graph.

\noindent\textbf{Are the multimodal experts redundant?}
Fig.~\ref{fig:ablation_modules}(b) further suggests that the multimodal design is not simply redundant with numerical temporal modeling. 
All modalities used by CausalMoE are deterministically transformed from the same raw time series, such as numerical patches, textual descriptions, and image-like renderings, rather than from external side information. 
Therefore, the improvement of the full model over \textit{w/o Multimodality} should be interpreted as a gain from richer representation learning under the same data budget. 
The semantic expert and the visual/multimodal expert also capture different aspects of the same temporal signal: the former encodes dataset context, statistical summaries, and task-oriented descriptions, while the latter emphasizes shape-level and structural patterns in the rendered time-series patches. 
This complementarity explains why removing multimodality weakens causal discovery, especially in noisy or data-scarce settings where raw numerical observations alone may induce spurious predictive links.

\noindent\textbf{Are the gains only due to large foundation backbones?}
Fig.~\ref{fig:pareto_frontier} answers this question by comparing accuracy and efficiency under different backbone choices. 
Although larger frozen LLM/VLM backbones further improve performance, the lightweight variant of CausalMoE still achieves strong causal discovery results while substantially reducing training cost. 
This demonstrates that the improvement does not merely come from parameter scaling, but from the proposed framework itself: pattern-routed heterogeneous experts provide specialized temporal representations, multimodal transformations regularize predictive dependency estimation, and CASA extracts sparse causal structures from the routed representations. 
In addition, since the foundation backbones are used as frozen feature extractors, their embeddings can be pre-computed and cached offline; online training and inference mainly involve the lightweight router, expert fusion, and CASA modules. 
Thus, CausalMoE provides a practical accuracy--efficiency trade-off: the full model offers the best causal discovery performance, while the lightweight variant preserves the key architectural advantages under limited computational budgets.

\section{Limitations and Ethical Considerations}
While CausalMoE achieves SOTA in few-shot causal discovery, we identify several limitations and ethical considerations.

\textbf{Limitations}. First, CausalMoE follows the GCD framework and therefore identifies predictive dependencies rather than interventional causality, relying on standard assumptions such as no hidden confounders and no instantaneous effects.
Second, although frozen foundation models enrich TS representations, they introduce additional embedding costs, which may limit use in strictly real-time scenarios.
Third, since the pre-training of LLMs and VLMs are not fully auditable, potential benchmark leakage cannot be completely ruled out; therefore, additional probing or counterfactual tests are recommended before applying into high-stakes GCD tasks.

\textbf{Ethical Considerations.} Users must not equate predictive GC with definitive physical mechanisms, especially in high-stakes domains
like healthcare. Moreover, reliance on pre-trained LLMs and VLMs risks inheriting
their biases or hallucinations, requiring rigorous validation before use in
sensitive decision-making.

\section{Conclusion}
In this paper, we introduce CausalMoE, the first billion-scale multimodal foundation model designed to overcome the limitations of Uniform Distribution Modeling in Granger causal discovery. By introducing a Pattern-Routed Mixture of Heterogeneous Experts, our framework explicitly models temporal heterogeneity and dynamically routes time-series patches to specialized domain experts to capture distinct mechanisms. Furthermore, we pioneered integrating LLM- and VLM-driven semantic priors into the causal discovery loop, using a causality-aware self-attention mechanism to infer interpretable, sparse graphs. Extensive evaluations on synthetic and real-world benchmarks demonstrate that CausalMoE significantly outperforms baselines. Notably, it exhibits exceptional robustness in few-shot scenarios, marking a paradigm shift towards regime-adaptive, data-efficient causal discovery.

\section{Acknowledgements}
This work is supported by the National Natural Science Foundation of China (62172018, 62102008), CCF-Tencent Rhino-Bird Open Research Fund (CCF-Tencent RAGR20250108), CCF-Zhipu Large Model Innovation Fund (CCF-Zhipu202414).



\bibliographystyle{ACM-Reference-Format}
\bibliography{main}

\end{document}